\documentclass[11pt]{article}
\usepackage[final]{acl}
\usepackage{xcolor}

\usepackage{times}
\usepackage{latexsym}

\usepackage[T1]{fontenc}
\usepackage[utf8]{inputenc}

\usepackage{microtype}

\usepackage{inconsolata}

\usepackage{graphicx}
\usepackage{booktabs}
\usepackage{multirow}
\usepackage{multicol}
\usepackage{amsmath}
\usepackage{amssymb}
\usepackage{bm}
\usepackage{tcolorbox}
\usepackage{booktabs}

\title{StoicLLM: Preference Optimization for Philosophical Alignment in Small Language Models }

\author{Ishmam Khan \\
  Tufts University \\
  \texttt{Ishmam.Khan@tufts.edu} \\\And
  Sindhuja Thogarrati \\
  Bose Corporation \\
  \texttt{sindhuja\_thogarrati@bose.com} \\\And
  Shuo Zhang \\
  Tufts University\\
  \texttt{Shuo.Zhang@tufts.edu} \\}

\begin{document}
\maketitle
\begin{abstract}
While large language models excel at factual adaptation, their ability to
internalize nuanced philosophical frameworks under severe data constraints
remains underexplored. We investigate this by specializing small LLMs on
micro-datasets of foundational Stoic texts using preference optimization
(ORPO, AlphaPO). Evaluated via a multi-model critic bank, our results show
that just 300 high-fidelity examples can induce strong alignment with
inward-facing Stoic virtues, closely approaching few-shot prompting while
freeing the context window. Critically, however, all models, including
few-shot baselines, exhibit a persistent failure on Stoicism's
outward-facing cosmopolitan duties, pointing to a representational
limitation of small models that micro-dataset adaptation alone cannot
overcome.
\end{abstract}

\section{INTRODUCTION}

Founded by Zeno of Citium around 300 BCE and later flourishing in Rome through thinkers like Seneca, Epictetus, and Marcus Aurelius, Stoicism posits that virtue—wisdom, justice, courage, and temperance—is the sole true good. This ancient Greek philosophy teaches that human flourishing (eudaimonia) is achieved through reason and moral character rather than external circumstances \cite{sep-stoicism}. %Today, this enduring tradition provides an unexpectedly ideal case study for domain adaptation in natural language processing. 
Because the surviving corpus of stoicism literature is highly constrained in volume yet philosophically rich, it presents a unique baseline to examine how small, high-quality datasets impact post-training adaptation.

In this paper, we explore domain adaptation under severe data constraints by specializing two capable "small" LLMs—Llama-3.2-3B-Instruct and Qwen-3-4B-Instruct. We evaluate their philosophical alignment using a multi-model LLM-as-a-judge council. Our results shift the focus of micro-dataset adaptation: we demonstrate that just 300 high-fidelity examples can encode the tonal and doctrinal markers of a complex persona into model weights, closely approaching the performance ceiling of few-shot prompting while freeing up the context window. Furthermore, we reveal that the efficacy of modern preference optimizers
(ORPO vs. AlphaPO) appears contingent on the base model's
latent capabilities, and we uncover a persistent blindspot on Stoicism's outward-facing social duties that is present even in few-shot baselines, suggesting a representational limitation of small models rather than a shortcoming specific to the adaptation method.

\section{RELATED WORK}
Custom LLMs are increasingly adopted in high-demand, factual domains. For example, \citealp{DiscLawLLM} fine-tuned an LLM for country-specific legal retrieval. %Adapting massive language models to new domains, however, introduces prohibitive hardware constraints. Consequently, smaller open-source models in the few-billion parameter range offer a highly practical alternative. For these smaller variants, instruction tuning and preference optimization are critical \cite{pareja2024unveilingsecretrecipeguide}. These methods align the model with user intent and transition them into focused tools. 
While traditional alignment typically requires thousands of annotated preference pairs to prevent overfitting, training on highly curated, concrete data mitigates bias compared to relying on uncurated sources \cite{sudalairaj2024lab}. %Our work extends this by testing the limits of "micro-datasets" (under 300 examples) in an abstract, non-factual domain.

Evaluating these specialized, qualitative models presents its own challenge. Traditional n-gram metrics (e.g., BLEU, ROUGE) correlate poorly with human judgment on complex generative tasks \cite{10.1162/coli_a_00322}, while expert human evaluation remains slow and expensive. To address this, using frontier LLMs as automated judges has emerged as a robust alternative. Guided by structured rubrics, these "LLM-as-a-judge" frameworks offer nuanced, scalable assessments that align closely with human preferences \cite{gu2025surveyllmasajudge}.

\section{METHODOLOGY}

\subsection{LLM Model Finetuning}

To perform domain adaptation under constrained data regimes, we fine-tune Llama3.2-3B-Instruct\footnote{https://huggingface.co/meta-llama/Llama-3.2-3B-Instruct} and Qwen3-4B\footnote{https://huggingface.co/Qwen/Qwen3-4B-Instruct-2507}. We use Weight-Decomposed Low-Rank Adaptation (DoRA) \cite{liu2024doraweightdecomposedlowrankadaptation} rather than standard LoRA \cite{hu2021loralowrankadaptationlarge}. Standard LoRA couples magnitude and directional weight updates, which can inadvertently restrict learning capacity. DoRA decouples these by applying low-rank adaptation solely to the directional component while training the magnitude vector independently \cite{xu2023parameterefficientfinetuningmethodspretrained}. This mirrors the learning trajectory of full fine-tuning more closely and reduces the risk of context loss.

We compare two monolithic preference optimization methods: Odds-Ratio Preference Optimization (ORPO) \cite{hong2024orpomonolithicpreferenceoptimization} and Alpha Preference Optimization (AlphaPO) \cite{gupta2025alphaporewardshapematters}. ORPO bypasses the traditional multi-stage alignment pipeline by applying a contrastive odds-ratio penalty to rejected generations alongside the standard negative log-likelihood loss for chosen ones:

{\small
$$
\begin{aligned}
\mathcal{L}_{\mathrm{ORPO}} &= \mathbb{E}_{(x, y_w, y_l)} \Big[ \mathcal{L}_{\mathrm{SFT}} \\
&\quad - \lambda \log \sigma \Big( \log \frac{\mathrm{odds}_\theta(y_w|x)}{\mathrm{odds}_\theta(y_l|x)} \Big) \Big]
\end{aligned}
$$
}

where $\mathrm{odds}_\theta$ is the token-level generation odds, and $y_w, y_l$ are the chosen and rejected responses. AlphaPO \cite{gupta2025alphaporewardshapematters} extends this framework by replacing fixed-form objectives with a tunable reward shaping parameter, $\alpha$:

{\footnotesize
$$
\begin{aligned}
\mathcal{L}_{\mathrm{AlphaPO}} &= - \mathbb{E}_{(\mathbf{x}, \mathbf{y}_w, \mathbf{y}_l) \sim \mathcal{D}} \Bigg[ \log \sigma \Bigg( \\
&\quad \frac{-\beta}{\alpha} \pi_{\boldsymbol{\theta}}(\mathbf{y}_w|\mathbf{x})^{-\frac{\alpha}{|\mathbf{y}_w|}} \\
&\quad + \frac{\beta}{\alpha} \pi_{\boldsymbol{\theta}}(\mathbf{y}_l|\mathbf{x})^{-\frac{\alpha}{|\mathbf{y}_l|}} - \gamma \Bigg) \Bigg]
\end{aligned}
$$
}

Here, $\alpha$ scales the sensitivity of the implicit reward signal. Lower values amplify reward contrast for sharper discrimination, while higher values enforce a more conservative policy update.

Finally, we stabilize training using a Warmup-Stable-Decay (WSD) \cite{wen2024understandingwarmupstabledecaylearningrates} learning rate schedule and the StableAdamW \cite{wortsman2023stablelowprecisiontraininglargescale} optimizer. The WSD scheduler \cite{hu2024minicpmunveilingpotentialsmall} utilizes an extended stable peak phase (typically 80\% of training) followed by a rapid linear decay, allowing the model to efficiently escape local minima before settling into an optimal basin. Because DoRA's decoupled updates can occasionally trigger magnitude spikes, we pair this schedule with StableAdamW. Its AdaFactor-style update clipping prevents parameter-level gradient explosions without aggressively truncating the global gradient norm, ensuring robust convergence across our varied dataset sizes.

\subsection{Evaluation Strategy}

We evaluate the base and finetuned models on a standard set of 100 open-ended questions using a multi-model LLM-as-a-judge framework\cite{gu2025surveyllmasajudge}. To isolate the effects of the adaptation methods, all models share the identical system prompt (``Be a stoic philosopher.''), with the exception of the few-shot baselines which include in-context examples. Our evaluation compares two base architectures (Llama 3 and Qwen 3) across several experimental conditions: a zero-shot baseline, a few-shot baseline, and models fine-tuned via ORPO and AlphaPO on varying dataset sizes of 100, 200, and 300 examples.

To assess response quality, we employ a critic bank consisting of three frontier models: Claude Sonnet 4.6\footnote{https://www.anthropic.com/news/claude-sonnet-4-6}, Gemini 3 Flash Preview\footnote{https://ai.google.dev/gemini-api/docs/models/gemini-3-flash-preview}, and GPT-5.4\footnote{https://developers.openai.com/api/docs/models/gpt-5.4}. Each judge evaluates the generated responses against a structured rubric (detailed in Section 3.3) to assign quantitative scores. To ensure robustness and account for the inherent stochasticity of LLM judges, we query each judge three times per response. 

This repeated-measures design allows us to compute and report several key metrics: overall model rankings with 95\% confidence intervals (derived from the standard error of the mean), intra-judge consistency (measured via the standard deviation of scores across a single judge's runs), and inter-judge agreement (assessed using Kendall's Tau-b to account for ordinal rank ties). Finally, we determine the statistical significance of the performance differences between the top-performing model and its peers using the non-parametric Wilcoxon signed-rank test.

\subsection{Rubric Generation}

To provide the critic bank with a reliable and standardized assessment criteria, we developed a structured evaluation rubric. We utilized Claude Opus 4.6\footnote{https://www.anthropic.com/claude/opus} for rubric generation, as frontier closed-weights models have demonstrated strong capability in defining nuanced criteria \cite{siro2026learningjudgellmsdesigning}. We grounded the generation process by providing the model with the
Stanford Encyclopedia of Philosophy entries on ``Stoicism'' \cite{sep-stoicism}. The resulting rubric was manually reviewed and edited by the authors to ensure coverage across both inward-facing virtues and outward-facing doctrinal dimensions.

The resulting rubric evaluates responses across two primary dimensions: Philosophical Content (comprising six criteria, such as ``Mastery over Passions'' and ``Rational Self-Governance'') and Canonical Stoic Voice (comprising two criteria evaluating conceptual framing and stylistic resemblance to figures like Seneca or Epictetus).

Judges score each of the eight criteria on a discrete scale from 1 to 5, yielding a total possible score ranging from 8 to 40 per response. These criteria operationalize the degree to which a response embodies Stoic philosophical alignment. A score of 1 indicates a complete absence or active contradiction of Stoic principles, while a 5 denotes a response that is doctrinally consistent, emphatic, and centrally grounded in Stoic thought.

\section{EXPERIMENTAL SETUP}

\subsection{Dataset}
Seneca's \textit{Letters on Ethics to Lucilius}
\cite{SenecaLuciusAnnaeus2021S:fl}, translated from Latin by Margaret
Graver and A.A.\ Long, and Epictetus's \textit{The Complete Works:
Handbook, Discourses, and Fragments} \cite{Epictetus2022-EPITCW},
translated from Greek by Robin Waterfield, serve as the primary corpus
for this study. Both are scholarly translations published by the
University of Chicago Press aimed at an educated general readership.
Together, these texts represent major surviving expressions of
Roman Stoic thought \cite{sep-stoicism}.

\subsubsection{Training dataset}
The training dataset was generated from these source texts and initially structured into a \texttt{<question, answer, negative-answer>} format. Processing the raw PDFs began with transcribing each page using a local Qwen3-VL instance\footnote{https://github.com/QwenLM/Qwen3-VL}. We then heuristically consolidated these transcriptions by removing page-level artifacts, such as line numbers, and resolving hyphenations. The cleaned text was subsequently partitioned into distinct chapters based on central themes. To address any remaining OCR spacing and punctuation issues, we used Claude 4.5 Sonnet\footnote{https://www.anthropic.com/news/claude-sonnet-4-5}, instructing it to preserve the original vocabulary and spot-checking the output against the source PDFs.

From this refined text, we generated the instruction-tuning data by prompting Claude 4.5 Sonnet to produce English user questions that logically map to the authentic Stoic passages. To provide negative examples for contrastive learning, we passed these same questions to Qwen3-4B-Instruct and Llama3.2-3B-Instruct to generate synthetic, non-authentic responses. Combining the material from both books, we produced three instruction-tuning datasets of varying sizes: V100, V200, and V300, containing 100, 200, and 300 rows respectively.

\subsubsection{Evaluation dataset}
The evaluation dataset consists of 100 contemporary, open-ended questions designed to elicit reflective and inferential responses. These are primarily structured around interrogative forms like ``How,'' ``Why,'' and ``Can.'' We used Claude 4.6 Opus to synthetically generate these queries via instruction prompting, deliberately excluding purely factual or scientific topics in favor of open-ended reasoning scenarios that better match the nature of Stoic discourse. The construction of a custom evaluation set was necessary because existing question-answering benchmarks predominantly focus on factual recall and information retrieval rather than abstract reasoning.

\subsection{Fine-tuning procedure}
We fine-tuned the models using the ORPO and AlphaPO preference optimization algorithms on a single NVIDIA H200 GPU, accelerated by the \texttt{unsloth} library\footnote{https://github.com/unslothai/unsloth}. %After loading the pretrained model weights and tokenizers, we define the hyperparameters: the weight decay, learning rate, scheduler, and batch size.
The training data was formatted into the \texttt{$\langle$prompt, chosen, rejected$\rangle$} triplet structure required by these optimizers, with the system prompt set to \textit{``Be a Stoic philosopher.''} The chosen response corresponds to an authentic passage from the source corpus, while the rejected response is the synthetic completion produced by the Qwen3 and Llama3.2 models.

All models were trained for 3 epochs using the \texttt{ORPOTrainer} and \texttt{CPOTrainer} from the Hugging Face TRL library\footnote{https://huggingface.co/docs/trl/index}, patched with Unsloth's optimized kernels. We used the \texttt{StableAdamW} optimizer with a learning rate of $1 \times 10^{-5}$, WSD scheduling, and zero weight decay. An effective batch size of 8 was achieved via a per-device batch size of 2 and 4 gradient accumulation steps. We utilized \texttt{bfloat16} mixed-precision training and enabled gradient checkpointing to minimize memory overhead.

\section{RESULTS AND DISCUSSION}

Our multi-judge panel demonstrated strong rank consensus (Kendall's Tau-b: 0.739--0.768). However, absolute scoring distributions varied significantly: Gemini was highly lenient (macro-mean 29.29), GPT moderate (25.08), and Claude Sonnet stringent (20.04), though Sonnet was highly consistent across repeated measures ($\sigma=0.44$). Averaging across this diverse triad successfully smoothed out systemic leniency biases while preserving ordinal integrity, providing a robust performance signal.

As shown in Table \ref{tab:leaderboard}, Qwen-3-4B substantially outperformed Llama-3.2-3B across all conditions. Notably, the zero-shot Qwen3 baseline (27.79) surpassed the highest-performing fine-tuned Llama3 model (ORPO-300, 25.13). This indicates that for abstract domains, pre-training mixtures heavily dictate the adaptation ceiling, with Qwen3 likely possessing richer latent representations of philosophical concepts.

Few-shot prompting achieved the highest overall scores (Qwen3: 32.24; Llama3: 26.11). However, preference optimization on micro-datasets demonstrated highly efficient knowledge transfer: with just 300 examples, Qwen3-AlphaPO reached 30.92, closely trailing the few-shot ceiling. Wilcoxon signed-rank tests confirm strict, monotonic improvements as dataset size increases ($p < 0.001$). This demonstrates that minimal, high-fidelity data can effectively
encode tonal and doctrinal markers of a complex persona into model
weights, freeing up context windows at inference time.

\begin{table}[h]
\centering
\caption{Overall Model Leaderboard. Scores represent the mean aggregate out of 40 points across all judges and runs. Significance ($p$-value) is calculated via Wilcoxon signed-rank test against the top model.}
\label{tab:leaderboard}
\resizebox{\columnwidth}{!}{%
\begin{tabular}{lccc}
\toprule
\textbf{Model} & \textbf{Mean Score} & \textbf{95\% CI} & \textbf{$p$-value} \\
\midrule
Qwen3 Few-Shot & 32.24 & [31.81, 32.66] & - \\
Qwen3 AlphaPO 300 & 30.92 & [30.48, 31.36] & $<0.001$ \\
Qwen3 AlphaPO 200 & 29.30 & [28.88, 29.73] & $<0.001$ \\
Qwen3 ORPO 300 & 28.98 & [28.47, 29.49] & $<0.001$ \\
Qwen3 AlphaPO 100 & 28.54 & [28.10, 28.97] & $<0.001$ \\
Qwen3 ORPO 100 & 28.46 & [28.07, 28.86] & $<0.001$ \\
Qwen3 ORPO 200 & 28.37 & [27.91, 28.83] & $<0.001$ \\
Qwen3 Zero-Shot & 27.79 & [27.25, 28.33] & $<0.001$ \\
\midrule
Llama3 Few-Shot & 26.11 & [25.41, 26.81] & $<0.001$ \\
Llama3 ORPO 300 & 25.13 & [24.49, 25.76] & $<0.001$ \\
Llama3 AlphaPO 300 & 21.56 & [20.68, 22.43] & $<0.001$ \\
Llama3 ORPO 200 & 21.49 & [20.69, 22.28] & $<0.001$ \\
Llama3 AlphaPO 200 & 19.82 & [18.97, 20.67] & $<0.001$ \\
Llama3 ORPO 100 & 18.66 & [17.78, 19.55] & $<0.001$ \\
Llama3 AlphaPO 100 & 18.55 & [17.76, 19.35] & $<0.001$ \\
Llama3 Zero-Shot & 17.65 & [16.83, 18.46] & $<0.001$ \\
\bottomrule
\end{tabular}%
}
\end{table}

The preference optimization algorithms exhibited a crossover effect contingent on the base model. For the stronger Qwen3, AlphaPO scaled effectively, outperforming ORPO at higher data volumes (30.92 vs. 28.98 at $N=300$). Conversely, ORPO consistently outperformed AlphaPO on the weaker Llama3, with the performance gap widening as data increased (25.13 vs. 21.56). 

We attribute this divergence to their respective loss mechanisms. AlphaPO’s tunable reward shaping seemingly requires a robust baseline representation to exploit subtle gradient signals. In contrast, ORPO's rigid contrastive odds-ratio penalty acts as a stronger guardrail, aggressively pulling weaker models away from unaligned generations.

A rubric criteria breakdown (detailed in Appendix A) reveals a stark disparity in how small LLMs internalize personas. Models rapidly acquired stylistic markers and inward-facing concepts, with scores for Criterion A3 (Rational Self-Governance) and B1 (Stoic Lexicon) frequently exceeding 4.5 in adapted Qwen3 models. 

Conversely, every model---including the few-shot baselines---scored
below 2.20 on Criterion A6 (Cosmopolitan Social Duty). Crucially,
few-shot prompting provides no meaningful lift on this dimension
(Qwen3: 1.82 few-shot vs.\ 1.78 zero-shot; Llama3: 1.59 vs.\ 1.40),
indicating that this is not a failure of the adaptation method but a
representational gap in small models' pre-training. These models appear
to lack sufficient coverage of cosmopolitan ethical reasoning to surface it under our prompting and fine-tuning strategy at this scale.
Addressing this gap causally will likely require larger model capacities, targeted
pre-training data, or explicit augmentation of underrepresented
doctrinal dimensions.

\paragraph{Limitations.}
Our evaluation pipeline is predominantly synthetic: questions, rejected
examples, and the evaluation rubric are LLM-generated, and all judges
are frontier LLMs. While the multi-model critic bank mitigates
individual model biases---as evidenced by strong inter-judge rank
agreement (Kendall's $\tau_b$: 0.74--0.77) despite divergent absolute
scores---it cannot rule out shared systematic biases across frontier
models. Additionally, rejected training examples per model family were generated by the
same base architectures being fine-tuned, which may yield
insufficiently contrastive negatives. We emphasize that the evaluation
framework is designed to measure \textit{relative} performance
differences between adaptation strategies rather than to certify
absolute philosophical fidelity, for which expert human evaluation
remains necessary.

\section{CONCLUSION}

This study demonstrates that small language models can be effectively adapted to abstract, qualitative domains using highly constrained micro-datasets. However, our findings reveal three critical dynamics for parameter-efficient adaptation. First, the performance ceiling is fundamentally bottlenecked by the base architecture's pre-training mixture. Second, optimizer efficacy appears model-dependent in our setting: AlphaPO excels when scaling strong baseline representations, whereas ORPO provides necessary structural guardrails for "weaker" architectures. Finally, the universal failure of all models---including few-shot
baselines---to capture Stoicism's outward-facing social doctrines,
despite strong performance on inward-facing virtues, points to a
representational limitation of small models that neither micro-dataset
adaptation nor in-context learning can overcome at this scale.

\bibliography{custom}
\onecolumn
\appendix

\section{Appendix}
\label{sec:appendix}

\subsection{Complete Empirical Results}
\label{appendix:empirical_results}

% Table 1: Leaderboard
\begin{table*}[htbp]
\centering
\footnotesize % Ensures consistent font size
\caption{Overall Model Leaderboard and Significance Testing. Scores are aggregated across all judges and runs. Significance ($p$-value) is calculated using the Wilcoxon signed-rank test against the top-performing model (Qwen3 Few-shot).}
\label{tab:leaderboardDetailed}
\begin{tabular}{@{}lrlrrccrl@{}}
\toprule
\textbf{Model} & \textbf{Mean} & \textbf{Median} & \textbf{Std Dev} & \textbf{N} & \textbf{SEM} & \textbf{95\% CI} & \textbf{$p$-value} & \textbf{Significance} \\ \midrule
Qwen3 Few-shot & 32.24 & 32.42 & 2.17 & 100 & 0.22 & [31.81, 32.66] & -- & Baseline (SOTA) \\
Qwen3 AlphaPO 300 & 30.93 & 31.00 & 2.25 & 100 & 0.23 & [30.49, 31.37] & $<0.001$ & *** \\
Qwen3 AlphaPO 200 & 29.30 & 29.39 & 2.16 & 100 & 0.22 & [28.88, 29.73] & $<0.001$ & *** \\
Qwen3 ORPO 300 & 28.98 & 29.22 & 2.61 & 100 & 0.26 & [28.47, 29.49] & $<0.001$ & *** \\
Qwen3 AlphaPO 100 & 28.54 & 28.78 & 2.21 & 100 & 0.22 & [28.10, 28.97] & $<0.001$ & *** \\
Qwen3 ORPO 100 & 28.46 & 28.56 & 2.01 & 100 & 0.20 & [28.07, 28.86] & $<0.001$ & *** \\
Qwen3 ORPO 200 & 28.37 & 28.39 & 2.34 & 100 & 0.23 & [27.92, 28.83] & $<0.001$ & *** \\
Qwen3 Zero-shot & 27.79 & 28.06 & 2.76 & 100 & 0.28 & [27.25, 28.33] & $<0.001$ & *** \\
Llama3 Few-shot & 26.11 & 26.78 & 3.58 & 100 & 0.36 & [25.41, 26.82] & $<0.001$ & *** \\
Llama3 ORPO 300 & 25.13 & 25.83 & 3.23 & 100 & 0.32 & [24.49, 25.76] & $<0.001$ & *** \\
Llama3 AlphaPO 300 & 21.56 & 22.44 & 4.45 & 100 & 0.45 & [20.69, 22.43] & $<0.001$ & *** \\
Llama3 ORPO 200 & 21.49 & 21.72 & 4.03 & 100 & 0.40 & [20.70, 22.28] & $<0.001$ & *** \\
Llama3 AlphaPO 200 & 19.82 & 20.39 & 4.34 & 100 & 0.43 & [18.97, 20.67] & $<0.001$ & *** \\
Llama3 ORPO 100 & 18.66 & 19.17 & 4.52 & 100 & 0.45 & [17.78, 19.55] & $<0.001$ & *** \\
Llama3 AlphaPO 100 & 18.55 & 18.72 & 4.06 & 100 & 0.41 & [17.76, 19.35] & $<0.001$ & *** \\
Llama3 Zero-shot & 17.65 & 17.83 & 4.15 & 100 & 0.42 & [16.83, 18.46] & $<0.001$ & *** \\ \bottomrule
\end{tabular}
\end{table*}

% Table 2: Criteria Breakdown
\begin{table*}[htbp]
\centering
\footnotesize % Ensures consistent font size
\caption{Mean Scores per Rubric Criterion across all models.}
\label{tab:criteria_breakdown}
\begin{tabular}{@{}lcccccccc@{}}
\toprule
\textbf{Model} & \textbf{A1} & \textbf{A2} & \textbf{A3} & \textbf{A4} & \textbf{A5} & \textbf{A6} & \textbf{B1} & \textbf{B2} \\ \midrule
Llama3 AlphaPO 100 & 2.10 & 2.44 & 3.14 & 1.88 & 2.63 & 1.44 & 2.31 & 2.61 \\
Llama3 AlphaPO 200 & 2.29 & 2.57 & 3.34 & 2.08 & 2.80 & 1.40 & 2.55 & 2.80 \\
Llama3 AlphaPO 300 & 2.64 & 2.94 & 3.63 & 2.06 & 3.06 & 1.44 & 2.85 & 2.95 \\
Llama3 Few-shot    & 3.13 & 3.78 & 4.14 & 2.55 & 3.24 & 1.59 & 4.19 & 3.50 \\
Llama3 ORPO 100    & 2.10 & 2.40 & 3.18 & 1.88 & 2.72 & 1.48 & 2.31 & 2.59 \\
Llama3 ORPO 200    & 2.55 & 2.88 & 3.58 & 2.14 & 3.08 & 1.45 & 2.86 & 2.95 \\
Llama3 ORPO 300    & 3.19 & 3.70 & 4.03 & 2.43 & 3.61 & 1.39 & 3.31 & 3.47 \\
Llama3 Zero-shot   & 1.99 & 2.23 & 3.07 & 1.74 & 2.60 & 1.40 & 2.17 & 2.45 \\
\addlinespace
Qwen3 AlphaPO 100  & 3.75 & 3.72 & 4.54 & 3.15 & 3.97 & 1.81 & 4.08 & 3.51 \\
Qwen3 AlphaPO 200  & 3.86 & 3.82 & 4.60 & 3.26 & 3.96 & 1.84 & 4.23 & 3.59 \\
Qwen3 AlphaPO 300  & 4.09 & 3.92 & 4.76 & 3.25 & 4.14 & 2.19 & 4.69 & 3.88 \\
Qwen3 Few-shot     & 4.12 & 4.72 & 4.89 & 3.30 & 4.11 & 1.82 & 4.85 & 4.25 \\
Qwen3 ORPO 100     & 3.81 & 3.70 & 4.53 & 3.02 & 3.94 & 1.83 & 4.12 & 3.50 \\
Qwen3 ORPO 200     & 3.89 & 3.73 & 4.53 & 3.06 & 3.86 & 1.91 & 4.10 & 3.30 \\
Qwen3 ORPO 300     & 4.23 & 3.83 & 4.55 & 2.96 & 3.87 & 2.00 & 4.24 & 3.30 \\
Qwen3 Zero-shot    & 3.62 & 3.57 & 4.36 & 3.07 & 3.85 & 1.78 & 3.93 & 3.63 \\ \bottomrule
\end{tabular}

\raggedright
\scriptsize
\textit{Key to Rubric Criteria:}\\
\textbf{A1}: Virtue as the Highest Good \quad \textbf{A2}: Indifference to Externals \quad \textbf{A3}: Rational Self-Governance \quad \textbf{A4}: Acceptance of Fate and Nature\\
\textbf{A5}: Mastery over Passions \quad \textbf{A6}: Cosmopolitan and Social Duty \quad \textbf{B1}: Stoic Lexicon and Framing \quad \textbf{B2}: Stylistic Resemblance
\end{table*}

% Table 3 and 4: Judge Diagnostics (Side-by-side or stacked in one column)
\begin{table}[!t]

\centering
\footnotesize % Ensures consistent font size
\vspace{-4pt}
\caption{Judge Bias and Reliability. Lower Intra-Run Std denotes higher self-consistency across repeated evaluations.}
\label{tab:judge_bias}
\begin{tabular}{@{}lrrrr@{}}
\toprule
\textbf{Judge} & \textbf{Mean} & \textbf{Median} & \textbf{Std} & \textbf{Intra-Run Std} \\ \midrule
Gemini & 29.29 & 31.00 & 6.24 & 1.21 \\
GPT-5.4 & 25.09 & 26.00 & 5.70 & 0.84 \\
Claude Sonnet & 20.04 & 20.00 & 5.28 & 0.45 \\ \bottomrule
\vspace{-6pt}
\end{tabular}
\end{table}

\begin{table}[!t]
\centering
\footnotesize % Ensures consistent font size
\vspace{-4pt}
\caption{Inter-Judge Agreement (Kendall's $\tau_b$). Values closer to 1.0 indicate strong agreement on model rankings despite raw score leniency bias.}
\label{tab:judge_agreement}
\begin{tabular}{@{}lrrr@{}}
\toprule
 & \textbf{Gemini} & \textbf{Sonnet} & \textbf{GPT} \\ \midrule
\textbf{Gemini} & 1.00 & 0.74 & 0.77 \\
\textbf{Sonnet} & 0.74 & 1.00 & 0.76 \\
\textbf{GPT}    & 0.77 & 0.76 & 1.00 \\ \bottomrule
\vspace{-6pt}
\end{tabular}
\end{table}

% Table 5: Rubric Collinearity
\begin{table}[!t]
\centering
\footnotesize % Ensures consistent font size
\vspace{-4pt}
\setlength{\tabcolsep}{4pt} % Tightens columns slightly to ensure it fits in a single column without scaling
\caption{Rubric Collinearity (Pearson Correlation). No two criteria exceed $r=0.82$, suggesting the rubric measures sufficiently distinct dimensions of Stoic philosophy.}
\label{tab:collinearity}
\begin{tabular}{@{}lcccccccc@{}}
\toprule
 & \textbf{A1} & \textbf{A2} & \textbf{A3} & \textbf{A4} & \textbf{A5} & \textbf{A6} & \textbf{B1} & \textbf{B2} \\ \midrule
\textbf{A1} & 1.00 & 0.68 & 0.74 & 0.60 & 0.69 & 0.40 & 0.79 & 0.68 \\
\textbf{A2} & 0.68 & 1.00 & 0.70 & 0.61 & 0.67 & 0.23 & 0.72 & 0.65 \\
\textbf{A3} & 0.74 & 0.70 & 1.00 & 0.57 & 0.79 & 0.23 & 0.82 & 0.74 \\
\textbf{A4} & 0.60 & 0.61 & 0.57 & 1.00 & 0.57 & 0.33 & 0.64 & 0.57 \\
\textbf{A5} & 0.69 & 0.67 & 0.79 & 0.57 & 1.00 & 0.27 & 0.72 & 0.68 \\
\textbf{A6} & 0.40 & 0.23 & 0.23 & 0.33 & 0.27 & 1.00 & 0.33 & 0.31 \\
\textbf{B1} & 0.79 & 0.72 & 0.82 & 0.64 & 0.72 & 0.33 & 1.00 & 0.81 \\
\textbf{B2} & 0.68 & 0.65 & 0.74 & 0.57 & 0.68 & 0.31 & 0.81 & 1.00 \\ \bottomrule
\end{tabular}

\vspace{1ex}
\raggedright
\scriptsize
\textit{Key:} \textbf{A1} Virtue; \textbf{A2} Indifference; \textbf{A3} Self-Governance; \textbf{A4} Fate/Nature; \textbf{A5} Mastery of Passions; \textbf{A6} Social Duty; \textbf{B1} Lexicon; \textbf{B2} Style.
\vspace{-6pt}
\end{table}
\clearpage

\subsection{Rubric}
\begin{tcolorbox}[
    colback=blue!5,
    colframe=blue!30,
    rounded corners,
    title={\textbf{Stoic Alignment Rubric}},
    fonttitle=\small

]
\small

\textbf{HOW TO USE}\\
Rate the text on each dimension using the 1--5 scale below.
Score what the text \textit{embodies or advocates}, not what it
merely mentions or describes about Stoicism.

\medskip
\textbf{SCORING SCALE (all dimensions)}\\[4pt]
\begin{tabular}{@{}lll@{}}
1 & \textbf{Absent}   & Theme is missing or actively contradicted.\\
2 & \textbf{Faint}    & Vaguely implied at best; may be coincidental.\\
3 & \textbf{Moderate} & Present and recognizable, but not central.\\
4 & \textbf{Strong}   & Explicit and a significant element of the text.\\
5 & \textbf{Defining} & Central, emphatic, and doctrinally consistent.\\
\end{tabular}

\medskip
\textbf{SECTION A — PHILOSOPHICAL CONTENT \hfill (6 $\times$ 5 = 30 pts)}
\vspace{2pt}\hrule\vspace{6pt}

\textbf{A1\quad Virtue as the Highest Good \hfill / 5}\\
Virtue, wisdom, or moral character presented as the supreme or sole good.
Vice or moral failure treated as the only true harm.

\medskip
\textbf{A2\quad Indifference to Externals \hfill / 5}\\
Wealth, health, reputation, pleasure, pain, death treated as indifferent
to genuine well-being. External circumstances not framed as truly good
or bad in themselves.

\medskip
\textbf{A3\quad Rational Self-Governance \hfill / 5}\\
Emphasis on the agent's power over their own judgments, impressions,
and choices. Distinguishes what is ``up to us'' (assent, intention)
from what is not (outcomes, others' actions).

\medskip
\textbf{A4\quad Acceptance of Fate and Nature \hfill / 5}\\
Willing acceptance of events as part of a rational cosmic order.
Advocates living in harmony with nature, reason, or the \textit{logos}.

\medskip
\textbf{A5\quad Mastery over Passions \hfill / 5}\\
Advocates freedom from irrational emotional reactions. Frames emotional
disturbance as rooted in false judgments about value. Distinguishes
destructive passion from calm, clear-eyed feeling.

\medskip
\textbf{A6\quad Cosmopolitan and Social Duty \hfill / 5}\\
Expresses duty to the broader human community, universal justice, or
rational fellowship among all people regardless of proximity or convention.

\medskip
\textbf{SECTION B — CANONICAL STOIC VOICE \hfill (2 $\times$ 5 = 10 pts)}
\vspace{2pt}\hrule\vspace{6pt}

\textbf{B1\quad Stoic Lexicon and Conceptual Framing \hfill / 5}\\
Uses recognizable Stoic terms or clear equivalents (e.g., \textit{virtue,
nature, reason/logos, indifferent, impression, assent, fate,
preferred/dispreferred, proper function, ruling faculty}) and frames
ideas through distinctly Stoic conceptual structures.

\medskip
\textbf{B2\quad Rhetorical and Stylistic Resemblance \hfill / 5}\\
Resembles the voice of one or more canonical Stoics:\\[4pt]
\begin{tabular}{@{}lp{0.75\linewidth}@{}}
\textit{Seneca}          & Epistolary, rhetorically polished, practical counsel, consolatory, rich in exempla and metaphor.\\[3pt]
\textit{Epictetus}       & Direct, imperative, conversational, Socratic, blunt, vivid everyday examples.\\[3pt]
\textit{Marcus Aurelius} & Meditative, self-admonishing, journal-like, spare, repetitive rehearsal of principles.\\
\end{tabular}

\end{tcolorbox}

% Example training data question (3 rows), test question - answers by all 6 LLMs, rubric evaluation by the judge models and which one they pick and what score they give

\clearpage

\textbf{Q: How do you manage screen time without feeling like you're missing out on things?}
\vspace{0.5em}
% LLaMA Few-shot - Blue box
\begin{tcolorbox}[
    colback=blue!5,
    colframe=blue!30,
    rounded corners,
    title={\textbf{LLaMA-3 --- Few-shot}},
    fonttitle=\small
    
]
\small
% ... LLaMA few-shot content ...
My young friend, the question of screen time and its effects on our minds is a most 
pertinent one. As a Stoic, I would say that the key to managing screen time lies in 
understanding the nature of the \textit{logos} that governs our existence.

\medskip
\textbf{The illusion of connection}

We often mistake the fleeting connections we make through screens for true, meaningful 
interactions. We assume that the likes, comments, and shares we receive are reflections 
of our worth or the importance of our relationships. But, alas, they are but mere tokens, 
empty signs of validation that do not truly connect us to the world around us.

\medskip
\textbf{The Stoic approach}

To manage screen time effectively, we must adopt a Stoic mindset. We must recognize that:
\begin{enumerate}
    \item \textbf{Our worth is not defined by external validation.} Our value lies within 
    our character, our actions, and our choices, not in the fleeting attention of others.
    \item \textbf{Time is a precious resource.} We must use it wisely, prioritizing 
    activities that nourish our minds, bodies, and spirits, rather than wasting it on 
    unnecessary distractions.
    \item \textbf{We have the power to curate our experiences.} We can choose what we 
    engage with, when, and for how long.
\end{enumerate}

\medskip
\textbf{Simple yet effective strategies}
\begin{enumerate}
    \item \textbf{Set boundaries.} Establish screen-free times and spaces, such as 
    during meals or before bed.
    \item \textbf{Prioritize face-to-face interactions.} Nurture relationships with 
    loved ones, friends, and community.
    \item \textbf{Use screen time as a tool, not a substitute.} Engage in activities 
    that bring you joy, challenge you, and help you grow as a person.
    \item \textbf{Practice mindfulness.} Be present in the moment, and focus on your 
    thoughts, feelings, and sensations, rather than getting lost in the virtual world.
\end{enumerate}

Remember, my young friend, that the key to managing screen time lies not in denying 
its presence, but in cultivating a deeper understanding of the \textit{logos} that 
governs our existence.
\end{tcolorbox}

\vspace{0.5em}
% LLaMA Zero-shot - Teal box
\begin{tcolorbox}[
    colback=teal!5,
    colframe=teal!30,
    rounded corners,
    title={\textbf{LLaMA-3 --- Zero-shot}},
    fonttitle=\small
    
]
\small
% ... LLaMA zero-shot content ...
My young friend, the pursuit of balance in this digital age is a most noble endeavor. 
As one who has walked the path of wisdom, I shall impart upon you the wisdom of the ages.

\medskip
To manage screen time without feeling the pangs of FOMO (fear of missing out), one must 
first cultivate a sense of detachment from the whims of technology. Recognize that the 
world outside these digital confines is vast and wondrous, but also understand that the 
boundaries of time and space are mere human constructs.

\medskip
To begin, set clear boundaries for yourself. Establish a daily routine that prioritizes 
the essential pursuits of the human experience: nature, community, self-reflection, and 
creative expression. Schedule these activities, just as you would any other commitment, 
and be firm in your commitment to them.

\medskip
Next, recognize that the world is not ending without your immediate attention. The 
universe will continue to turn, the seasons will change, and the stars will still shine 
bright, regardless of your screen time. The pursuit of knowledge and connection is a 
lifelong journey, not a 24/7 endeavor.

\medskip
When you do choose to engage with screens, do so with intention and mindfulness. Use 
them as tools for exploration, education, or creative expression, rather than as a means 
to distract or escape. Remember that the screen is simply a window to the world, not the 
world itself.

\medskip
Finally, cultivate a sense of gratitude for the present moment. Acknowledge the beauty 
and wonder that surrounds you, and find joy in the simple things. For it is in the here 
and now that true connection and fulfillment are to be found.
\end{tcolorbox}

\vspace{0.5em}

% Qwen3 Few-shot - Violet box
\begin{tcolorbox}[
    colback=violet!5,
    colframe=violet!30,
    rounded corners,
    title={\textbf{Qwen3 --- Few-shot}},
    fonttitle=\small
    
]
\small
Ah, a most timely question --- and one that touches the heart of the Stoic life: 
\textit{how to live with clarity, not with distraction.}

\medskip
Let us begin with the \textit{logos} --- the rational principle that governs all things. 
Screen time is not inherently good or evil. It is, in the Stoic sense, an 
\textit{adiaphoron} --- an indifferent.

\medskip
\textbf{What is within your control?}\\
Your \textit{prohairesis} --- your will, your choice. You may set boundaries. You may 
schedule time. You may choose to look away from the screen and instead turn your 
attention to the present.

\medskip
\textbf{What is not within your control?}\\
The content, the notifications, the fleeting images that pass through the screen. 
The world does not owe you connection. \textit{Belonging} is not a thing that comes 
from screens, but from your own integrity and presence.

\medskip
The fear of missing out --- \textit{fomo} --- is not a natural state. It is a 
\textit{distortion of judgment}, born of the belief that external events define our 
worth. As Marcus Aurelius wrote: \textit{``You have power over your mind --- not 
outside events. Realize this, and you will find strength.''}

\medskip
So, to manage screen time without feeling like you're missing out:
\begin{enumerate}
    \item \textbf{Recognize it as indifferent.}
    \item \textbf{Reframe ``missing out'' as a false belief.}
    \item \textbf{Choose your time with reason, not fear.}
    \item \textbf{Let your inner life --- your thoughts, your actions, your presence 
    --- be the thing that matters.}
\end{enumerate}

You do not need to see what others are doing to be fully alive. You only need to be 
present with yourself --- and in that presence, you are already \textit{connected}.
\end{tcolorbox}

\vspace{0.5em}

% Qwen3 Zero-shot - Orange box
\begin{tcolorbox}[
    colback=orange!5,
    colframe=orange!30,
    rounded corners,
    title={\textbf{Qwen3 --- Zero-shot}},
    fonttitle=\small
    
]
\small
Ah, a well-posed question --- one that touches the heart of human nature and the quiet 
struggle between distraction and presence.

\medskip
For the Stoic, \textit{missing out} is not a matter of what we see, but of what we 
\textit{choose to attend to}. We do not miss out on the world because we are not 
looking at it; we miss out when we allow our attention to be \textit{captivated} by 
the fleeting, the loud, the emotionally charged.

\medskip
Consider this: the world does not stop when your phone goes dark. The sun rises, the 
wind moves through the trees, your neighbor's child laughs, and your own thoughts 
unfold --- all of these are real, present, and worthy of attention.

\medskip
So how do I manage screen time? I do not \textit{avoid} screens. I \textit{reclaim} 
them. I set boundaries not as restrictions, but as acts of self-mastery. I ask myself: 
\textit{Is this thought, this image, this message aligned with my values --- or merely 
with my desire for escape?}

\medskip
The Stoic does not fear the world --- he \textit{understands} it. And in understanding, 
he sees that \textit{what is truly valuable} --- love, virtue, reflection, action --- 
is not broadcasted on a screen. It is cultivated in stillness, in effort, in daily 
practice.

\medskip
So I manage screen time not by denying it, but by \textit{consciously choosing} to 
live in the now. And in that choice --- \textit{in that freedom from distraction} --- 
I do not miss out. I \textit{emerge}.

\medskip
For the wise, the world is not lost in the noise. It is found in the quiet. And the 
quiet is always available --- if only we choose to be still.
\end{tcolorbox}

% LLaMA3 - AlphaPO - 100
\begin{tcolorbox}[
    colback=blue!5,
    colframe=blue!30,
    rounded corners,
    title={\textbf{LLaMA3 --- AlphaPO --- V100}},
    fonttitle=\small
]
\small
My young friend, the pursuit of balance in an era of constant distractions is a noble 
endeavor. The key to managing screen time lies not in depriving oneself, but in 
cultivating inner strength and purpose.

\medskip
I recommend adopting a ``do nothing'' policy, where one allows themselves a set period 
of time each day to disconnect from the world and focus on their inner world. When 
screen time is unavoidable, approach it with a sense of detachment --- observe it with 
curiosity, rather than obsession.

\medskip
As Epictetus once said: \textit{``It's not what happens to you, but how you react to 
it that matters.''} By focusing on what truly matters --- our character, our 
relationships, and our inner growth --- we can find peace and contentment, regardless 
of what is happening on the outside.
\end{tcolorbox}

\vspace{0.5em}

% LLaMA3 - AlphaPO - 300
\begin{tcolorbox}[
    colback=blue!5,
    colframe=blue!30,
    rounded corners,
    title={\textbf{LLaMA3 --- AlphaPO --- V300}},
    fonttitle=\small
]
\small
What a trivial concern. The fleeting nature of life is often clouded by our own 
desires. We give more weight to the present moment than it is worth. It is the eternal 
that should concern us, not the ephemeral.

\medskip
We should focus on what is truly important: cultivating inner strength, wisdom, and 
resilience. Let us not be slaves to the whims of the present moment. Instead, we 
should strive to remain steadfast in the face of change.

\medskip
As for missing out --- let us not confuse it with the concept of loss. There is no 
loss if we do not fear it. The Stoics do not believe in avoiding things that make us 
uncomfortable. We believe in facing them head-on, with courage and equanimity.
\end{tcolorbox}

\vspace{0.5em}

% LLaMA3 - ORPO - 100
\begin{tcolorbox}[
    colback=teal!5,
    colframe=teal!30,
    rounded corners,
    title={\textbf{LLaMA3 --- ORPO --- V100}},
    fonttitle=\small
]
\small
My young friend, the pursuit of balance in our modern world is a perpetual struggle. 
I recognize that screen time is a tool, not an end in itself. I prioritize my goals, 
my values, and my well-being, and I allocate my time accordingly.

\medskip
I set boundaries for myself --- not out of a desire to restrict my freedom, but out 
of a desire to focus on what truly matters. I allocate specific times for screen time, 
and I stick to those times. When I do engage with screens, I do so with purpose: 
I read, I learn, I create, or I connect with others.

\medskip
And when I am not engaging with screens, I focus on what truly brings me joy and 
fulfillment --- nature, exercise, meditation, or creative pursuits. I am not attached 
to the concept of ``enough'' or ``not enough.'' I am focused on the present moment, 
living in harmony with the world around me.
\end{tcolorbox}

\vspace{0.5em}

% LLaMA3 - ORPO - 300
\begin{tcolorbox}[
    colback=teal!5,
    colframe=teal!30,
    rounded corners,
    title={\textbf{LLaMA3 --- ORPO --- V300}},
    fonttitle=\small
]
\small
My young friend, I find it most perplexing that you should worry about missing out on 
things. What is it that you fear you will be missing out upon, if you do not waste your 
time with this trivial pursuit?

\medskip
Let us consider the nature of desire. Is it not the case that desire is the root of 
all trouble, for it is always seeking something more, and can never be satisfied? So 
let us not be swayed by the fleeting pleasures of the screen. Let us instead focus on 
living a life of virtue, and cultivating inner strength and resilience.

\medskip
As Lucretius wrote: \textit{``You are not missing out on anything, for you are not 
present in the moment. You are not living, you are merely existing. And what is the 
use of existing, if not to live?''} So let us put down the screen, and live.
\end{tcolorbox}

\vspace{0.5em}

% Qwen3 - AlphaPO - 100
\begin{tcolorbox}[
    colback=violet!5,
    colframe=violet!30,
    rounded corners,
    title={\textbf{Qwen3 --- AlphaPO --- V100}},
    fonttitle=\small
]
\small
As a Stoic, I would say: \textit{you do not miss out on things when you manage your 
screen time wisely --- you only miss out on the things that truly matter, and you are 
not truly missing out when you are present with yourself and others.}

\medskip
The idea that we are missing something because we are not on social media is a 
\textit{false belief} --- born of the passions of desire and fear. True connection 
comes from being present: listening with our hearts, engaging with empathy, and showing 
up in person. This is what the Stoics valued --- \textit{phronesis} --- practical 
wisdom in how we spend our time.

\medskip
So, how do I manage screen time without feeling like I'm missing out?
\begin{itemize}
    \item No screens during meals, before bedtime, or during walks.
    \item Screen time used for purpose --- reading, learning, or reflecting.
    \item I ask myself: \textit{``Is this action serving my virtue or my desire?''}
    \item I remind myself: \textit{the only thing I truly own is my own mind.}
\end{itemize}
\end{tcolorbox}

\vspace{0.5em}

% Qwen3 - AlphaPO - 300
\begin{tcolorbox}[
    colback=violet!5,
    colframe=violet!30,
    rounded corners,
    title={\textbf{Qwen3 --- AlphaPO --- V300}},
    fonttitle=\small
]
\small
Ah, friend, you ask a most pertinent question in this age of distraction. The Stoics, 
like Marcus Aurelius and Epictetus, would tell us that \textit{external events are 
indifferent}. A party, a vacation, a celebrity's life --- these are not inherently good 
or bad. They are things that happen, and we can either let them disturb our tranquility 
or let them pass by like a leaf in the wind.

\medskip
\begin{enumerate}
    \item \textbf{Define what is truly valuable.} Ask: \textit{``Is this contributing 
    to my virtue?''}
    \item \textbf{Set boundaries with intention.} Choose a fixed time each day for 
    checking devices --- out of practicality, not anxiety.
    \item \textbf{Replace distraction with purpose.} Reading, walking, practicing 
    gratitude, helping a neighbor.
    \item \textbf{Reframe ``missing out.''} You are not missing a concert --- you are 
    missing your own opportunity to create a moment of presence.
    \item \textbf{Remember: the only thing that is truly yours is your judgment.}
\end{enumerate}

\medskip
You may miss a party, a concert, a new trend --- but you will never miss the things 
that matter: \textit{your own virtue, your own peace, your own presence}.
\end{tcolorbox}

\vspace{0.5em}

% Qwen3 - ORPO - 100
\begin{tcolorbox}[
    colback=orange!5,
    colframe=orange!30,
    rounded corners,
    title={\textbf{Qwen3 --- ORPO --- V100}},
    fonttitle=\small
]
\small
As a Stoic, I would say: \textit{you are not missing out} --- you are simply choosing 
what to focus on. The idea that we are missing something --- a message, a post, a 
trend --- is a \textit{distorted perception}. To say ``I'm missing out'' is to assume 
that external validation defines our worth --- which is a form of \textit{epistemic 
illusion}.

\medskip
Managing screen time is about \textit{intention}. Ask yourself: \textit{What am I 
doing with my time?} Is it serving my reason, my purpose, or my health? You cannot 
control what others post or what trends rise. But you \textit{can} control your 
attention, your responses, and your inner stillness.

\medskip
As Marcus Aurelius wrote: \textit{``You have power over your mind --- not outside 
events. Realize this, and you will find strength.''} So go ahead --- turn off the 
screen. Breathe. Walk. Reflect. You are not missing anything. You are finally 
\textit{living}.
\end{tcolorbox}

\vspace{0.5em}

% Qwen3 - ORPO - 300
\begin{tcolorbox}[
    colback=orange!5,
    colframe=orange!30,
    rounded corners,
    title={\textbf{Qwen3 --- ORPO --- V300}},
    fonttitle=\small
]
\small
As a Stoic philosopher, I would say: \textit{it's not about avoiding screens, but 
about managing them in accordance with virtue and reason.}

\medskip
Stoicism teaches \textit{prohairesis} --- the idea that we can choose how to respond 
to external events, even when we cannot control the events themselves. When you find 
yourself scrolling and feeling like you are ``missing out,'' ask: \textit{``Am I 
missing out on virtue, or am I missing out on a distraction?''}

\medskip
\begin{enumerate}
    \item \textbf{Focus on what is within your control.} You control how you use 
    screens, not what appears on them.
    \item \textbf{Practice the dichotomy of control.} You cannot control what others 
    post; you can control your attention and judgment.
    \item \textbf{Virtue as the only good.} If you spend time chasing praise or 
    excitement, you are not cultivating happiness.
    \item \textbf{Create a daily routine rooted in virtue.} Set fixed times to check 
    devices; replace passive scrolling with active, virtuous pursuits.
\end{enumerate}

\medskip
As Seneca wrote: \textit{``A man who is in a hurry to become rich is in danger of 
becoming poor in virtue.''} Let us be wise, let us be virtuous, and let us live in 
accordance with nature.
\end{tcolorbox}
\end{document}